\documentclass[letterpaper]{article} 
\usepackage{aaai25}  
\usepackage{times}  
\usepackage{helvet}  
\usepackage{courier}  
\usepackage[hyphens]{url}  
\usepackage{graphicx} 
\usepackage{natbib}  
\usepackage{caption} 
\usepackage{multirow}
\usepackage{booktabs}
\usepackage{amsmath,amsfonts,amsthm,amssymb}
\usepackage{algorithm}
\usepackage{algorithmic}

\usepackage{newfloat}
\usepackage{listings}
\DeclareCaptionStyle{ruled}{labelfont=normalfont,labelsep=colon,strut=off} 
\floatstyle{ruled}
\newfloat{listing}{tb}{lst}{}
\floatname{listing}{Listing}
\title{\textit{CustomContrast}: A Multilevel  Contrastive Perspective For Subject-Driven Text-to-Image Customization}
\author {
    Nan Chen,
    Mengqi Huang,
    Zhuowei Chen,
    Yang Zheng,
    Lei Zhang,
    Zhendong Mao\thanks{Corresponding author}
}
\affiliations {
     University of Science and Technology of China\\
    \{chen\_nan, huangmq, chenzw01, zy849900389\}@mail.ustc.edu.cn, \{leizh23, zdmao\}@ustc.edu.cn
}

\usepackage{bibentry}

\begin{document}

\maketitle

\begin{abstract}
Subject-driven text-to-image (T2I) customization has drawn significant interest in academia and industry. This task enables pre-trained models to generate novel images based on unique subjects. Existing studies adopt a self-reconstructive perspective, focusing on capturing all details of a single image, which will misconstrue the specific image's irrelevant attributes (\emph{e.g.}, view, pose, and background) as the subject intrinsic attributes. This misconstruction leads to both overfitting or underfitting of irrelevant and intrinsic attributes of the subject, \emph{i.e.}, these attributes are over-represented or under-represented simultaneously, causing a trade-off between similarity and controllability. In this study, we argue an ideal subject representation can be achieved by a cross-differential perspective, \emph{i.e.}, decoupling subject intrinsic attributes from irrelevant attributes via contrastive learning,  which allows the model to focus more on intrinsic attributes through intra-consistency  (features of the same subject are spatially closer) and inter-distinctiveness (features of different subjects have distinguished differences). Specifically, we propose \textbf{\textit{CustomContrast}}, a novel framework, which includes a Multilevel Contrastive Learning (MCL) paradigm and a Multimodal Feature Injection (MFI) Encoder. The MCL paradigm is used to extract intrinsic features of subjects from high-level semantics to low-level appearance through crossmodal semantic contrastive learning and multiscale appearance contrastive learning.  To facilitate contrastive learning, we introduce the MFI encoder to capture cross-modal representations. Extensive experiments show the effectiveness of \textit{CustomContrast} in subject similarity and text controllability.

\end{abstract}

\begin{links}
     \link{code}https://cn-makers.github.io/CustomContrast/
\end{links}

%

\section{Introduction}

Subject-driven text-to-image (T2I) customization aims to empower pre-trained text-to-image models to generate images of unique subjects (\emph{e.g.}, objects, animals) specified by users. Given an image of a specific subject, users can generate novel depictions of the subject in various scenes, poses, and motions guided by target prompts. This task has drawn rapidly growing research interests in both academia and industry because of its broad applications. The key challenge of customized T2I generation lies in how to precisely generate the target subject while maximally preserving the text controllability of the pre-trained diffusion models.



\begin{figure}[!t]
\centering
\includegraphics[width=1\linewidth]{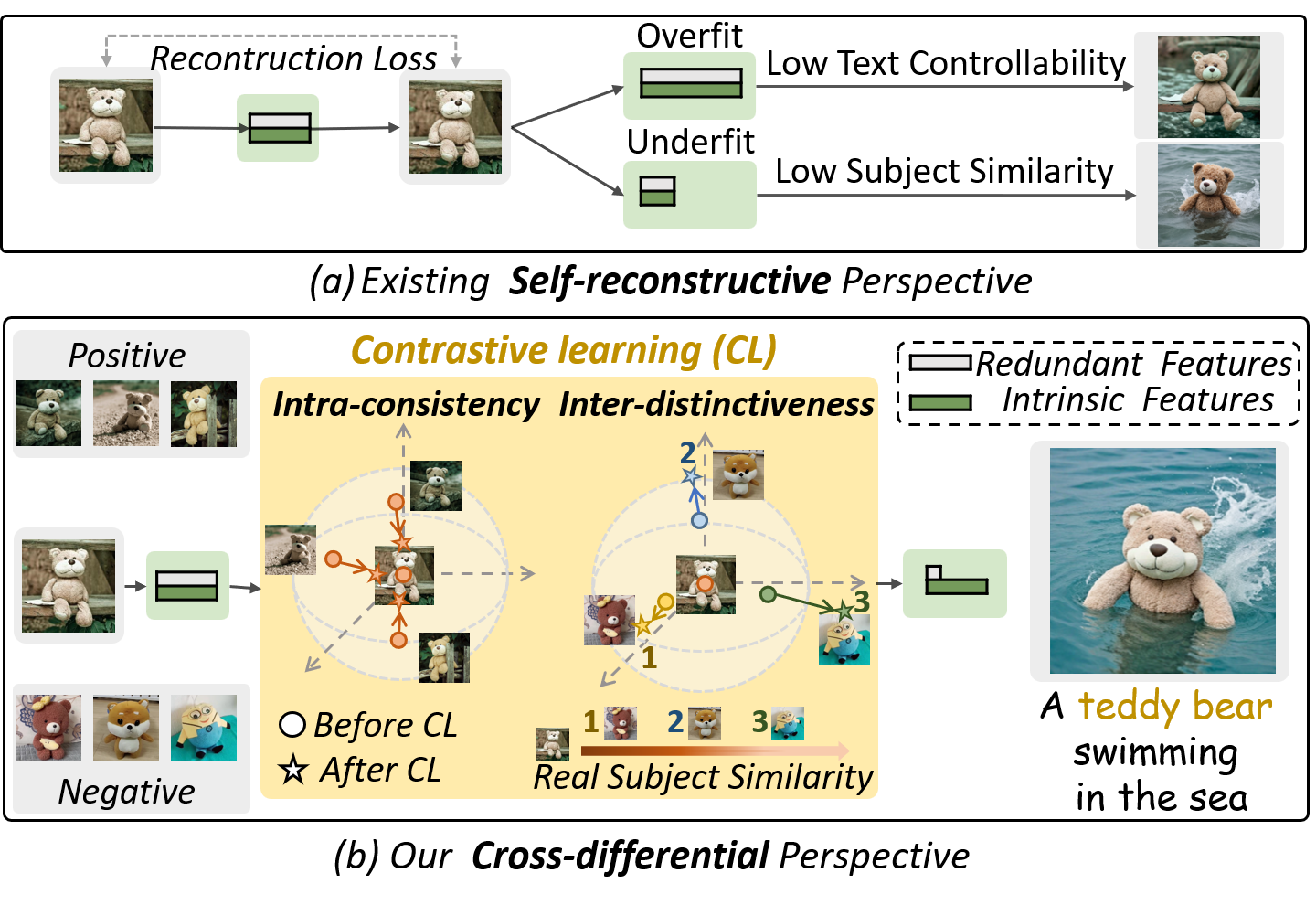}
\caption{Comparison with existing perspective. (a) Existing studies learn each subject feature with entangled redundant features (\emph{e.g.}, view, pose), suffering a trade-off between similarity and controllability (redundant and intrinsic features simultaneously overfit or underfit since they are coupled together). (b) In contrast, we rethink it from a cross-differential perspective.  By using contrastive learning to ensure intra-consistency (features of the same subject are spatially closer) and inter-distinctiveness (features of different subjects have distinguished differences), our model disentangles the subject intrinsic features from irrelevant features for dual optimization of controllability and similarity.}
\label{fig_intro}
\end{figure}

Existing customized T2I methods mainly involve two streams: the finetune-based stream and the finetune-free stream. The finetune-based stream \cite{gal2022image,ruiz2023dreambooth,kumari2022customdiffusion} typically fine-tunes each new concept.  In this paper, we focus on the finetune-free stream, which achieves real-time generation by training adapters based on the pre-trained encoders. Early work \cite{wei2023elite, ye2023ip} aims to enhance subject similarity by injecting image embeddings into U-Net using additional visual attention. Recent studies further improve text controllability while preserving subject similarity by selectively injecting image features. Some work \cite{li2024blip, zhang2024ssr} reduces background interferences by using different reference and reconstruction images, while other work \cite{song2024moma} controls the generated regions through additional inference operations. Additionally, some work \cite{sun2024generative} introduces multimodal large language models to further enhance text controllability. To summarize, existing methods adopt a self-reconstructive perspective, that is, taking target subject images as inputs, aiming to fully mimic all the subject details via the diffusion reconstruction loss \cite{ho2020denoising}.

However, the self-reconstructive perspective has inherent defects, \emph{i.e.}, misconstructing the attributes in a specific image as the intrinsic attributes of the subject. The image's attributes are entangled, including irrelevant attributes (\emph{e.g.}, orientation, pose, and background), and intrinsic attributes of the subject that remain unchanged despite external interferences.  Thus, reconstruction from a single image of the subject will inevitably result in the capture of both irrelevant and subject-intrinsic attributes. As a result, Insufficient representations result in inaccurate extraction of both intrinsic and irrelevant attributes, reducing similarity (\emph{e.g.}, underfitting in Fig. \ref{fig_intro}(a) 
with changes in the toy's color). Conversely, over-representations cause overfitting of both intrinsic and irrelevant attributes, reducing controllability (\emph{e.g.}, overfitting in Fig. \ref{fig_intro}(a) with the toy's pose and scene hard to alter). Expanding the dataset or changing the image background still follows this perspective, therefore these cannot significantly improve the misconstruction of redundant attributes.

 Therefore, we argue an ideal subject representation can be achieved by a \textbf{cross-differential} perspective, \emph{i.e.}, comparing differences between target samples via contrastive learning, which aims to capture each subject's accurate representation. 
 As shown in Fig. \ref{fig_intro}(b), this perspective achieves intra-consistency and inter-distinctiveness. Firstly, Intra-consistency is achieved by pulling images of the same subject under different contexts closer, decoupling irrelevant attributes. Secondly, Inter-distinctiveness is ensured by comparing the specific subject with others, thereby further learning the fine-grained intrinsic features. These allow the model to focus more on the intrinsic attributes than redundant attributes. By reducing redundant attributes' interference, the cross-differential perspective can achieve more accurate extraction of subject intrinsic features while enabling flexible interaction with text (\emph{e.g.}, the swimming toy in Fig. \ref{fig_intro}(b)).

 In this paper, we propose a novel customization framework, \textbf{\textit{CustomContrast}}, from the cross-differential perspective, which uses contrastive learning to extract intrinsic attributes for dual optimization of controllability and similarity.
\textit{CustomContrast}  includes two key components: the  Multimodal Feature Injection (MFI) Encoder, which aims to generate consistent multimodal representations to support the implementation of contrastive learning, and a Multilevel Contrastive Learning (MCL) paradigm that focuses on contrasting subject representations from low-level appearance to high-level semantics. Specifically, the MFI-Encoder includes Visual Qformer, Textual Qformer, and TextualVisual (TV) Fusion module. The first two modules are used to extract visual and textual embeddings. The TV Fusion module uses textual embeddings as queries to capture text-related visual embeddings. The novel MCL paradigm includes Crossmodal Semantic Contrastive Learning (CSCL) and Multiscale Appearance Contrastive Learning (MACL). CSCL enhances the semantic consistency between learned textual and visual embeddings, while MACL preserves the relative distances of samples between the learned embedding space and the real subject space. 

Our main contributions are summarized as follows:
\begin{itemize}
    \item \textbf{Concepts.}  We proposes \textit{CustomContrast}, a novel paradigm from a cross-differential perspective. By ensuring intra-consistency and inter-distinctiveness, it gradually extracts the intrinsic subject representations.
    \item  \textbf{Technology.}  We propose a novel MCL paradigm that extracts intrinsic representations of subjects from high-level semantics to low-level appearance through CSCL and MACL. To support the MCL paradigm, we introduce the MFI-encoder to capture cross-modal representations.
    \item  \textbf{Experiments.} Our model, trained on SD-V1.5 and SDXL, outperforms corresponding advanced methods. Experiments show our model improves text controllability by 3.8\% and 5.4\% respectively, and subject similarity (E-DI) by 5.9\% and 2.4\%, while easily extending to multi-subject and human domain generation.
\end{itemize}


\begin{figure*}[!t]
\centering
\includegraphics[width=1\linewidth]{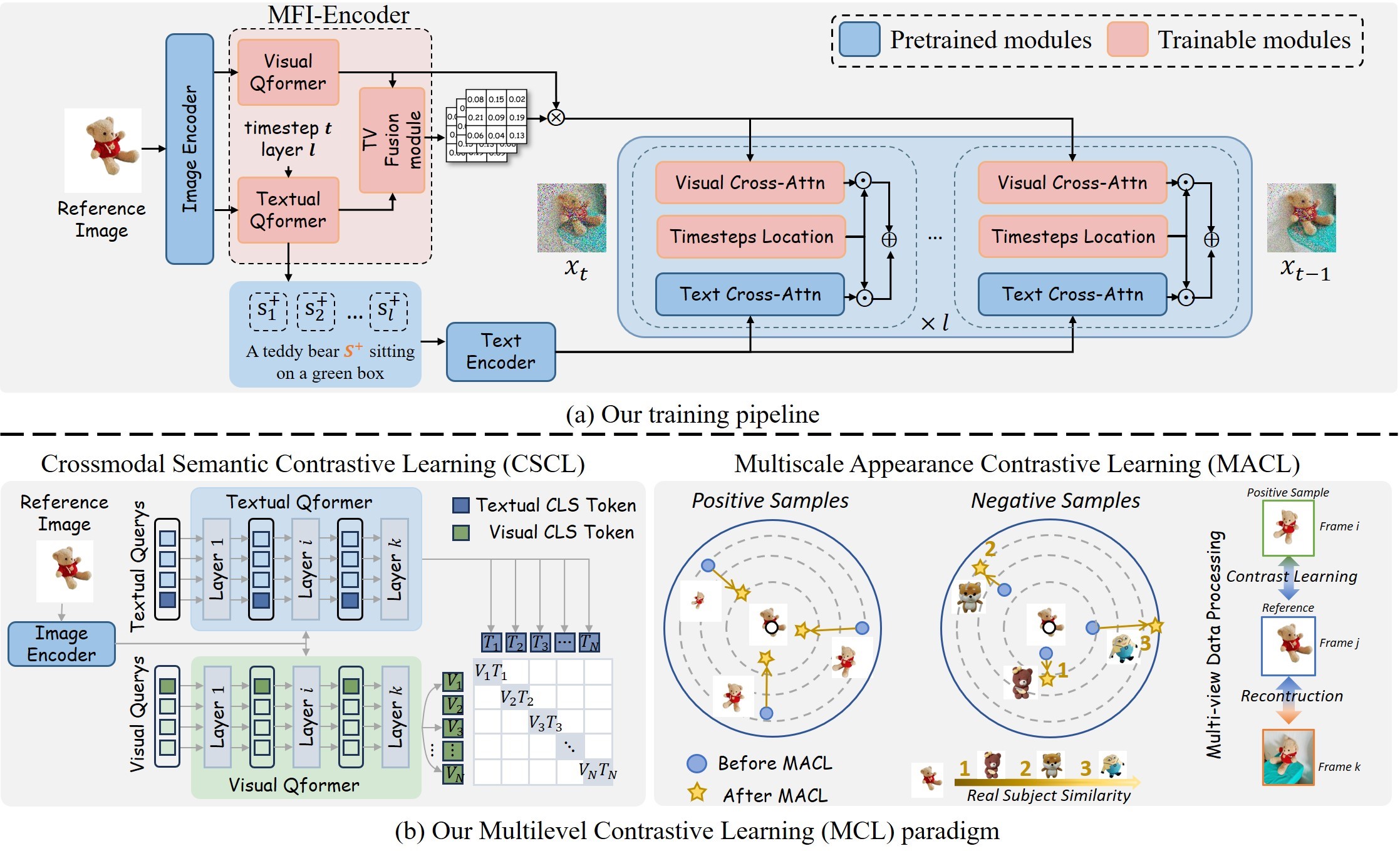}
\caption{Overview of the proposed  \textit{CustomContrast}. (a) Training pipeline. The consistency between textual and visual features is accurately learned by the MFI-Encoder, which includes a Textual-Visual (TV) Fusion module to enhance feature consistency from visual and textual Qformers. (b) The MCL paradigm includes CSCL, aligning high-level semantics by contrasting visual and textual embeddings via \texttt{CLS} tokens,  and MACL, 
which is applied to text embeddings from different cross-attention layers. MACL decouples redundant subject features by aligning positive samples (segmented images of the same subject from various views, positions, and sizes), while preserving relative distances by contrasting with other subjects.} 
\label{fig_model_architecture}
\end{figure*}


\section{Related Work}
\subsection{Subject-Driven Text-to-image Customization}
Existing customized text-to-image work includes finetune-based streams and finetune-free streams. Early work  \cite{gal2022image,ruiz2023dreambooth,kumari2022customdiffusion,han2023svdiff,2023p+,alaluf2023neural,zhang2023prospect,nam2024dreammatcher,huang2022dse} focuses on the finetune-based stream, which requires learning pseudo-words as the subject representations and fine-tuning each new subject. Custom Diffusion  \cite{kumari2022customdiffusion} and SVDiff  \cite{han2023svdiff} reduce computational overhead by minimizing fine-tuning parameters. Recent work   \cite{2023p+,zhang2023prospect} in the $ \mathcal{P}^+$ space uses different pseudo-words for different cross-attention and timesteps.

The finetune-free stream aims to eliminate optimization for each subject, achieving real-time generation. Recent work  \cite{wei2023elite,li2024blip,huang2024realcustom,zhang2024ssr,song2024moma,kosmos-g,wang2024ms,patel2024lambda,sun2024generative,chen2023dreamidentity,huang2023not,huang2023towards}  trains an additional adapter to map reference images to image prompts. Elite  \cite{wei2023elite} uses local mapping to enhance subject features. IP-adapter \cite{ye2023ip} injects visual features into additional cross-attention, struggling with attribute changes (\emph{e.g.}, shape and orientation). RealCustom  \cite{huang2024realcustom,mao2024realcustom++} and SSR-Encoder \cite{zhang2024ssr} selectively inject subject-related features to reduce background features.  Other studies \cite{song2024moma,kosmos-g} use multimodal large language models for image-text alignment. 





\subsection{Contrastive Learning of Representations}

Early zero-shot classification studies  \cite{khattak2023maple, CoPrompt} based on CLIP \cite{radford2021learning} have validated the effectiveness of contrastive learning in aligning visual features with textual features. By dynamically adjusting these representation spaces in downstream tasks, the consistency between visual and textual representations can be further improved. In customized tasks, past work based on the Multimodal Large Language Model (MLLM) \cite{patel2024lambda,kosmos-g,sun2024generative} has explored using contrastive learning to reduce the gap between image and text spaces. Summarily,  similar to traditional contrastive learning, previous studies in customized task primarily focus on aligning image embeddings with text embeddings,  usually at the expense of subject similarity. In contrast, our work uses contrastive learning to extract intrinsic features from redundant features. It decouples redundant features through intra-consistency and further learns intrinsic features through inter-distinctiveness, achieving dual optimization of text controllability and subject similarity.




\section{Methodology}

The pipeline of \textit{CustomContrast} is depicted in Fig. \ref{fig_model_architecture}, which mainly contains  MFI-Encoder and Multilevel Contrastive Learning (MCL) paradigm.  Given reference images during training, \textit{CustomContrast} extracts consistent multimodal representations through the MFI-Encoder, which is designed to support the implementation of contrastive learning. Then, the MCL paradigm contrasts these representations from low-level appearance to high-level semantics, extracting intrinsic subject representations. In this section, we will first introduce the preliminaries of the diffusion model. Then, we will describe the MFI-Encoder and MCL paradigm in detail.

\subsection{Preliminaries}



Our work is based on Stable Diffusion (SD) \cite{rombach2022high}, which includes an autoencoder and a UNet denoiser. The encoder $\mathcal{E}(\cdot)$ of the autoencoder maps a given image $\boldsymbol{x}\in \mathbb{R}^{H \times W\times 3}$ to a low-dimensional space $\boldsymbol{z}=\mathcal{E}(\boldsymbol{x})\in \mathbb{R}^{h \times w\times d}$, where $d$ denotes  the hidden space dimension. The corresponding decoder $\mathcal{D}(\cdot)$ maps $\boldsymbol{z}$ back to pixel space.  In customized task, the  text \( \boldsymbol{c}_t \) and the reference image \( \boldsymbol{c}_i \) are used as conditional inputs to the denoiser \( \epsilon_\theta(\cdot) \).
Ultimately, the denoiser $\epsilon_\theta(\cdot)$  is trained by mean-squared loss: 

\begin{equation}
 \mathcal{L}_{\text{LDM}}=\underset{ \substack{ \boldsymbol{z},   \boldsymbol{\epsilon}, t}}{ \mathbb{E}} \left\|\boldsymbol{\epsilon}-\epsilon_\theta\left(\boldsymbol{z}_t, t,\boldsymbol{c}_t,\boldsymbol{c}_i \right)\right\|_2^2,\label{s}
\end{equation}
where $\boldsymbol{\epsilon}$ refers to  unscaled noise and $t$ means denoising timestep. $\boldsymbol{z}_t$ is the hidden space tensor at $t$-th timestep.  

\subsection{Multimodal Feature Injection Encoder}

The Multimodal Feature Injection (MFI) Encoder is designed to extract multimodal features (textual and image features), which consists of three main components: Visual-Qformer, Textual-Qformer, and the TV Fusion Module. 

The  CLIP image encoder is used to extract image features. Previous studies \cite{wei2023elite,zhang2024ssr}  show that using only the last layer's feature of CLIP loses fine-grained details. We propose to concatenate features from different CLIP layers to get coarse-to-fine subject features \( \boldsymbol{f}_I = \left\{ \boldsymbol{f}_k \right\}_{k=0}^K \), where \( K \) refers to layer numbers, defaulting to 3. Then \( \boldsymbol{f}_I \) is fed into textual and visual Qformer.


Here, Qformer denotes the Transformer module where the queries are learnable queries, and image features are input as part of keys and values. PerceiverAttention (PA)  \cite{awadalla2023openflamingo} is used in each Qformer to aggregate image features with learnable queries. \( PA(\boldsymbol{f}_I, \boldsymbol{f}_q) \) aggregates image features \( \boldsymbol{f} _I\) and learnable queries \( \boldsymbol{f}_q \)  as follows:
\begin{equation}
\label{eq:pa}
 \text{Softmax}\left( \frac{ \boldsymbol{Q}( \boldsymbol{f}_q)  \boldsymbol{K}\left( [ \boldsymbol{f} _I, \boldsymbol{f}_q] \right)}{\sqrt{d}} \right)  \boldsymbol{V}\left( [ \boldsymbol{f} _I,  \boldsymbol{f}_q] \right),
\end{equation}
where \( [\boldsymbol{f} _I, \boldsymbol{f}_q] \) denotes concatenation of features \( \boldsymbol{f} _I \) and \( \boldsymbol{f}_q \).

To enable MFI Encoder to dynamically adapt to variations in timesteps and UNet layers  $l$, we use Fourier mapping and learnable embeddings for UNet layer and timestep information to generate spatiotemporal queries \( \boldsymbol{f}_{q}^{st} \in  \mathbb{R}^{(l+1) \times d} \). These queries \( \boldsymbol{f}_{q}^{st} \) are used by Textual Qformer to query the image features \( \boldsymbol{f}_I \), mapping subjects to predefined text space as \( \boldsymbol{f}_t \in  \mathbb{R}^{(l+1) \times d} \). The first $l$ features of \( \boldsymbol{f}_t \)  are respectively fed into the \( l \) text cross-attention layers after passing through the text encoder. Differently, vanilla learnable visual queries \( f_q^v \) are used by the visual Qformer to extract visual features  \( \boldsymbol{f}_v \in  \mathbb{R}^{(m+1) \times d}\), where $m+1$ denotes visual query numbers. The first  $m$ features of \( \boldsymbol{f}_v\) are finally injected into each visual cross-attention layer. The remaining single features of \( \boldsymbol{f}_t \) and \( \boldsymbol{f}_v \) are used for CSCL in Section 3.3. 


%



\subsubsection{TV Fusion Module}

To further align visual features \( \boldsymbol{f}_v \)  with text space, we use the  \( PA(\boldsymbol{f}_t, \boldsymbol{f}_v) \) module in Eq. \ref{eq:pa} to extract refined visual features \( \boldsymbol{\hat{f}}_{v} \) related to text features \( \boldsymbol{f}_t \). Finally, \( \boldsymbol{\hat{f}}_{v} \) and \( \boldsymbol{f}_t \) are injected into parallel visual and text cross-attention. The attention integration is as follows:
\begin{equation}
    \begin{aligned}
    & \quad \quad \text{Attention}(\boldsymbol{Q}, \boldsymbol{K}_t, \boldsymbol{V}_t, \boldsymbol{K}_v, \boldsymbol{V}_v)= \\
      & \text{Softmax}\left(\frac{\boldsymbol{Q}\boldsymbol{K}_t^\top}{\sqrt{d}}\right)\boldsymbol{V}_t + \gamma \cdot\text{Softmax}\left(\frac{\boldsymbol{Q}\boldsymbol{K}_v^\top}{\sqrt{d}}\right)\boldsymbol{V}_v,
    \end{aligned}  
\end{equation}
where $\boldsymbol{Q} =  \boldsymbol{W}_q \boldsymbol{z}_t$, $\boldsymbol{K}_v = \boldsymbol{W}^v_k \boldsymbol{\hat{f}}_{v}$, $\boldsymbol{K}_t = \boldsymbol{W}^t_k \boldsymbol{\hat{f}}_{t}$, $\boldsymbol{V}_v = \boldsymbol{W}^v_v \boldsymbol{\hat{f}}_{v}$, and $\boldsymbol{V}_t = \boldsymbol{W}^t_v \boldsymbol{\hat{f}}_{t}$. $\boldsymbol{W}_q$, $\boldsymbol{W}_k^t$, $\boldsymbol{W}_v^t$ are frozen, while $\boldsymbol{W}_k^v$ and $\boldsymbol{W}_v^v$ are trainable projection layers for visual cross attention. $\boldsymbol{z}_t$ denotes image latents and $\boldsymbol{\hat{f}}_{t}$ denotes text feature generated by feeding $\boldsymbol{f}_t$ and the corresponding caption into the text encoder. $\gamma$ is a weight factor, defaulting to 1.


\subsection{Multilevel Contrastive Learning Paradigm}



 Cross-differential perspective aims to capture subjects' distinguished intrinsic features by contrasting  
 multilevel differences between subjects. To achieve this, we propose a Multilevel Contrastive Learning paradigm: Crossmodal Semantic Contrastive Learning aligns the textual and visual features for semantic intrinsic consistency, and Multiscale Appearance Contrastive Learning ensures feature distances are consistent with those of real subjects across different scales.


\subsubsection{Crossmodal Semantic Contrastive Learning}
\label{sec:ti_section}

Features of subjects in the CLIP text and image space are extracted through Visual and Textual Qformer, respectively. However,  a gap often exists between these features, making it challenging for  CLIP image features to respond sensitively to complex transformations (\emph{e.g.}, shape transformations).

To address this,   we propose Crossmodal Semantic Contrastive Learning (CSCL) to align the features generated by the Visual Qformer and Textual Qformer, ensuring they are in the same space, as shown in Fig. \ref{fig_model_architecture}(b)(left). We add a visual \texttt{CLS} token $\boldsymbol{v}^{cls}$ and a textual \texttt{CLS} token $\boldsymbol{t}^{cls}$  to the visual queries and textual queries, applying CSCL loss $\mathcal{L}_{c}$ on them, which is as follows:
\begin{equation}
\begin{aligned}
\mathcal{L}_{c}=&-\sum_{i \in N }\log \left[\frac{\exp \left( a_{ii} \right)}{\sum_{j \neq \mathrm{i}} \exp \left(a_{ji} \right)}\right]
\\&-\sum_{i \in N }\log \left[\frac{\exp \left( a_{ii} \right)}{\sum_{j \neq \mathrm{i}} \exp \left(a_{ij} \right)}\right],
\end{aligned} 
\end{equation}
where   \(i\) and \(j\) donate two different samples within a batch,  $N$ donates batch size.  \(a_{ij}\) represents  \( \cos(\boldsymbol{f}^{cls}_{v_i}, \boldsymbol{f}^{cls}_{t_j}) \),  measuring  consistency between  \(\boldsymbol{f}^{cls}_{vi}\) and \(\boldsymbol{f}^{cls}_{t_j}\)  by cosine similarity. $\boldsymbol{f}^{cls}_{v_i}$ refers to the remaining \texttt{CLS} features of i-th sample's $\boldsymbol{f}_{v}$, and $\boldsymbol{f}^{cls}_{t_j}$ refers to the remaining \texttt{CLS} feature of j-th sample's $\boldsymbol{f}_{t}$.

%


\subsubsection{Multiscale Appearance Contrastive  Learning}
The key idea of Multi-scale Appearance Similarity Contrastive Learning (MACL) is to ensure that the distance relationships between multiscale features are consistent with those of real subjects.  This means the features of the same subject with different situations should be as close as possible (intra-consistency), while the distances between different samples' features should match those between real subjects (inter-distinctiveness). As shown in Fig.  \ref{fig_model_architecture}(b)(right), we achieve intra-consistency by pulling positive samples of the reference subject closer, and inter-distinctiveness by introducing scaling factors to align the feature distances with negative samples to real subject distances. In this section, we will introduce the $\textit{S}^+$ Space and  MACL  in the $\textit{S}^+$  Space.

\subsubsection{A. $\textit{S}^+$  Space}\label{s_space}
\ 

\begin{figure}[!t]
\centering
\includegraphics[width=1\linewidth]{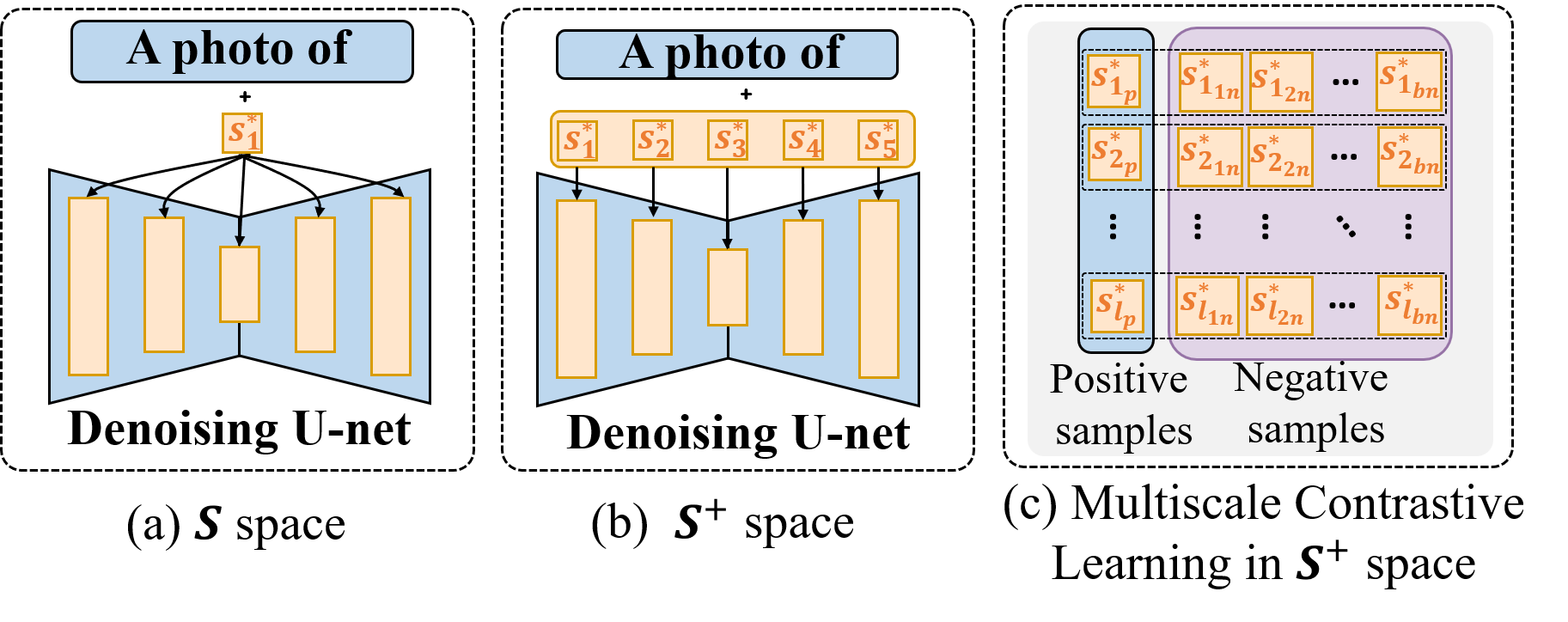}
\caption{(a)  In $\textit{S}$ space, a  token $\boldsymbol{s^*}$ influences all cross-attention layers. (b)  In  $ \textit{S}^+$ space, different $\boldsymbol{s^*_i}$ control cross-attention layers. (c) MACL is applied separately to each $\boldsymbol{s^*_i}$.}
\label{fig_ss}
\end{figure}

\indent Injecting token $\boldsymbol{s}^* \in \mathbb{R}^{k \times d} $ into  textual space is used for finetune-based studies \cite{ruiz2023dreambooth} as shown in Fig. \ref{fig_ss}(a). Here,  $k$ and $d$ refer to token numbers and dimensions.  Define $ \textit{S}$ as the space spanned by $\boldsymbol{s}^*$. We then define extended visual concept space $ \textit{S}^+$ as  follows:
\begin{equation}
  \textit{S}^+:=[\boldsymbol{s}^*_1,\boldsymbol{s}^*_2,\cdots,\boldsymbol{s}^*_l],
\end{equation}
where $\boldsymbol{s}^*_i$ denotes  tokens for $i$-th  cross-attention layer and $l$ denotes  cross-attention layer numbers.  $\boldsymbol{s}^+  \in \mathbb{R}^{l \times k \times d} $ is a tensor in the $\textit{S}^+$ space. As depicted in Fig. \ref{fig_ss}(b),  \(l\) features of \(\boldsymbol{s}^+\) are injected into  \(l\) cross-attention. MACL is applied to   \(\boldsymbol{s}_i^*\)  for different cross-attention as shown in Fig. \ref{fig_ss}(c).


\subsubsection{B. MACL in $\textit{S}^+$ Space}
\ 

\indent MACL  in the  $\textit{S}^+$ space preserves the multi-scale similarity structure, ensuring the similarities of learned features are positively correlated with those of real subjects. Thus, the sample similarities must satisfy the following conditions:
\begin{equation}
\mathcal{D}_r(\boldsymbol{x}_i,\boldsymbol{x}_j)  < \mathcal{D}_r(\boldsymbol{x}_i,\boldsymbol{x}_k)  \Leftrightarrow   \mathcal{D}_g(\boldsymbol{z}_i,\boldsymbol{z}_j)  < \mathcal{D}_g(\boldsymbol{z}_i,\boldsymbol{z}_k),
\label{condition}
\end{equation}
where $\mathcal{D}_r(\cdot)$  denotes the  ideal distance metric for  real subjects $\boldsymbol{x}$, and $\mathcal{D}_g(\cdot)$  is  relative distance metric for learned features $\boldsymbol{z}$. $i$, $j$ and $k$  denote the $i$-th,  $j$-th,  and $k$-th samples.  We use the CLIP image encoder \(\psi_C(\cdot)\) to extract appearance features of real subjects, approximating relative distance $\mathcal{D}_r$  between real subjects. Note that \(\psi_C(\cdot)\) is a specific case to approximate $\mathcal{D}_r$ and can be replaced by others (\emph{e.g.}, DINO-V2 \cite{sinhamahapatra2024finding}, SigLIP \cite{zhai2023sigmoid}).

We design  MACL  scaling factors to implement the aforementioned constraints. These factors scale the generated samples' similarity based on the real subjects' similarity. We use cosine similarity, denoted as $cos(\cdot)$,  to measure the similarity between the $\boldsymbol{s^+}$   components of different samples at all $l$ cross-attention layers.  The appearance representation of the segmented subject images $\boldsymbol{\hat{x}}$, obtained by CLIP image encoder, is utilized to compute the appearance similarity matrix $R_{ap}=\{r_{ij}\}_{i,j=1}^{N \times N}$, where $r_{ij}=cos(\psi_C(\boldsymbol{ \hat{x}}_i),\psi_C(\boldsymbol{\hat{x}}_j))$. Here, segmented subject images $\boldsymbol{\hat{x}}$   approximate the real subjects $\boldsymbol{x}$, allowing MACL to focus on the subjects themselves rather than the background. The appearance scaling factor is $\alpha_{ij}=1/r_{ij}$. The MACL loss $\mathcal{L}_{m}^{ij}$  between  $\boldsymbol{s^+_i}$ and $\boldsymbol{s^+_j}$ extracted by Textual Qformer  is:
\begin{equation}
\begin{aligned}
\mathcal{L}_{m}^{ij}=
-\sum_{a=0}^{l}  \log \left[\frac{\exp \left(\alpha_{i j}\cdot cos_{i j}^a  / \tau\right)}{\sum_{k \neq \mathrm{i}}^{N} \exp \left(\alpha_{i k} \cdot cos_{i k}^a) / \tau\right)}\right],
\end{aligned} 
\end{equation}
where  $cos_{i j}^a$  and $cos_{i k}^a$  are abbreviation for $cos(\boldsymbol{s}^+_{ia},\boldsymbol{s}^+_{ja})$  and  $cos(\boldsymbol{s}^+_{ia},\boldsymbol{s}^+_{ka})$.  $\boldsymbol{s}^+_{ia}$, $\boldsymbol{s}^+_{ja}$ and $\boldsymbol{s}^+_{ka}$ respectively represent  components  of the $a$-th cross attention layer for $i$-th sample's $\boldsymbol{s}^+_{i}$,  $j$-th sample's $\boldsymbol{s}^+_{j}$, and   $k$-th sample's $\boldsymbol{s}^+_{k}$, $\tau$ denotes  temperature and $N$ dnotes batch size. The scaling factor  $\alpha_{ij}$  aligns relative distance relationships between learned features and real subjects. The detailed derivation is in Appendix A.1.

As shown in Fig. \ref{fig_model_architecture}(b) (right), We select frames different from the reference images as MACL positive samples. By aligning images of the same subject, \textit{CustomContrast}  effectively decouples irrelevant features of the subject. The processing details of positive samples are in Appendix B.


\subsection{Timesteps-specific Subject Location}


\begin{figure}[!t]
\centering
\includegraphics[width=1\linewidth]{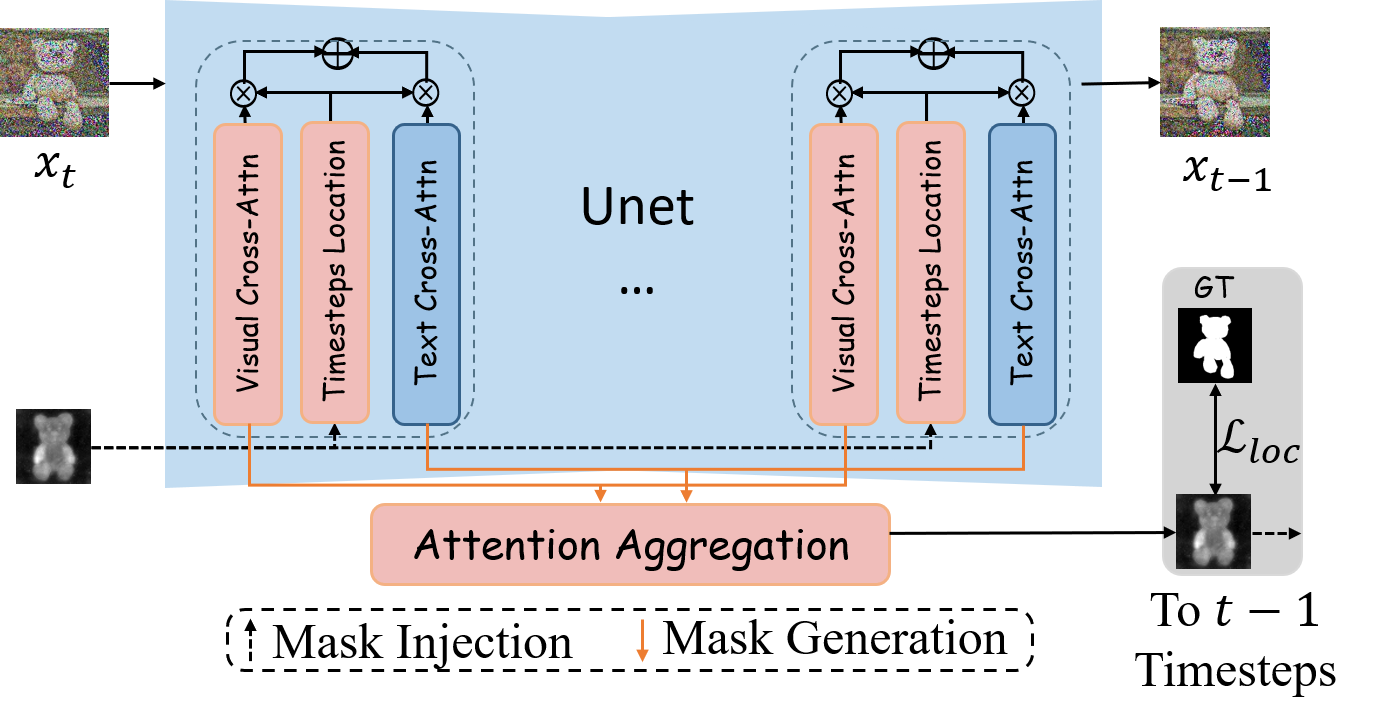}
\caption{Illustration of timesteps-specific subject location.}
\label{fig_mask_generation}
\end{figure}

  
Similar to MOMA\cite{song2024moma}, we adopt timestep injection inference operations. The main difference is that we use location loss \cite{xiao2023fastcomposer} to further reduce redundant features' interference, as shown in Fig. \ref{fig_mask_generation}.
During training, we extract all visual cross-attention maps \(A_v\) and p-th text cross-attention maps \(A_t^p\)  related to \(\mathcal{S}^+\) space (where \(p\) is the placeholder index). The global attention map \(\overline{A}_g\) is then aggregated by averaging these attention maps. \(\overline{A}_g\) is constrained with the actual mask $ M_g$ by  loss function $\mathcal{L}_{\text{loc}}$ \cite{xiao2023fastcomposer}:
\begin{equation}
    \mathcal{L}_{\text{loc}} = \text{mean} \left( (1 - M_g) \odot (1-\overline{A}_g) \right) - \text{mean} \left( M_g \odot\overline{A}_g) \right),
\end{equation}
where $ \odot$ denotes the element-wise multiplication. During inference, the extracted mask at timestep \(t+1\) is injected into the visual attention maps and the p-th text attention maps at timestep \(t\), as shown in  Fig. \ref{fig_mask_generation}. Then, the location mask generated at timestep \(t\) is injected in the next timestep \(t-1\), and so on. The visualization details are in Appendix C.7. 

The total loss is as follows:
\begin{equation}
    \mathcal{L}_{\text{total}} = \mathcal{L}_{\text{LDM}} + \lambda_1 (\mathcal{L}_{\text{c}} + \mathcal{L}_{\text{m}}) + \lambda_2 (\mathcal{L}_{\text{loc}}),
\end{equation}
where $\lambda_1$ and $\lambda_2$ are loss hyper-parameters.

\section{Experiments}
\subsection{Experimental Setups}

\textbf{Implementation.} Our model is implemented on both SD V1-5 and SDXL. The filtered subset of MVImageNet \cite{yu2023mvimgnet} and OpenImage \cite{kuznetsova2020open} are used as training sets. The model is trained on 6 A100 GPUs for 200k iterations with learning rate 3e-5 and $\lambda_1$=1e-2 and $\lambda_2$=1e-3.  The layer numbers of Textual and Visual Qformer are set to 4.  More details are in Appendix C.1.

\textbf{Test Prompt.} To evaluate complex scenarios, we create a prompt set with four categories: style, shape, scene, and accessorization modification. See Appendix C.2 for details.

\textbf{Evaluation metrics.}  We evaluate our model based on DINO-I (DI)  \cite{oquab2023dinov2}, CLIP-I (CI), and CLIP-T (CT)  scores on DreamBench \cite{ruiz2023dreambooth}. \textit{\textbf{Similarity.}} To exclude background interference, we calculate the subject similarity after segmenting the reference subjects and generated subjects with Grounded SAM \cite{ren2024grounded}. Similarity metrics are refined by measuring subject similarity in reconstruction (R-CI and R-DI) and editing (E-CI and E-DI).   \textit{\textbf{Controllability.}} CLIP-T metric is calculated using cosine similarity in the CLIP text-image embedding space.  Additionally, ImageReward (IR) \cite{xu2024imagereward} is also used to evaluate controllability.

\begin{table}[!t]
    \centering
   
    {
    \fontsize{7}{8.4}\selectfont
        \begin{tabular}{c c cc cccc}
       \toprule
         \multirow{2}{*}{ Methods } & \multicolumn{2}{c}{ \textbf{\textit{controllability}} }   &\multicolumn{4}{c}{\textbf{ \textit{similarity }}} \\
         \cmidrule(r){ 2 - 3 }  \cmidrule(r){ 4- 7 } & CT $\uparrow$ & IR $\uparrow$ & R-CI $\uparrow$ & E-CI $\uparrow$ & R-DI $\uparrow$ & E-DI $\uparrow$ \\
        \midrule Custom Diffusion  & 0.258  & -1.41 & 0.863 & 0.756 & 0.661  & 0.517 \\
        DreamMatcher  & 0.298 & 0.16 & 0.875  & 0.781   & 0.670 & 0.540 \\
        \midrule  BLIP-Diffusion  & 0.278& -0.80  &  0.865& 0.768  & 0.629 &  0.535 \\ 
        ELITE  & 0.291& -0.66 &  0.856& 0.762 & 0.661 & 0.533 \\
         IP-Adapter-plus  & 0.295& -0.61 &  0.878 &  0.766 & 0.702 & 0.558 \\
         SSR-Encoder  &  0.303& -0.89 & 0.857  &  0.767 & 0.623 & 0.524 \\
         KOSMOS-G  & 0.268 & -1.45 &  0.864 & 0.753   & 0.677 &  0.531\\
         MOMA  & 0.313 & -0.24 & 0.879  & 0.772   & 0.679 &  0.550\\

        \midrule \textbf{ours (SD-v1.5)}   & \textbf{0.325} & \textbf{0.45} & \textbf{0.919} & \textbf{0.788} & \textbf{0.737} & \textbf{0.591}  \\
        \bottomrule

        \end{tabular}
    }
     \caption{Quantitative results with SD-v1.5 based methods. Bolded numbers indicate the best performance.}
    \label{Quantitative_results_1}
\end{table}

\begin{table}[!t]
    \centering
   
    {
    \fontsize{7}{8.4}\selectfont
        \begin{tabular}{ccc cccc}
            \toprule
            \multirow{2}{*}{Methods} & \multicolumn{2}{c}{\textbf{\textit{controllability}}} & \multicolumn{4}{c}{\textbf{\textit{similarity}}} \\
            \cmidrule(r){2-3} \cmidrule(r){4-7}
            & CT $\uparrow$ & IR $\uparrow$ & R-CI $\uparrow$ & E-CI $\uparrow$ & R-DI $\uparrow$ & E-DI $\uparrow$ \\
            \midrule
            $\lambda$-ECLIPSE   & 0.279  & -1.03  & 0.882  & 0.773  & 0.677  & 0.511 \\
            IP-Adapter-XL   & 0.289 & -0.71 & 0.918 & 0.790 & 0.723 & 0.570 \\
            EMU2    & 0.298 &  -0.203 & 0.91 & 0.795 & 0.719 &  0.583 \\
            MS-Diffusion   & 0.312 &  -0.24 & 0.901 & 0.772 & 0.682 & 0.540 \\
            \midrule
            \textbf{ours (SDXL)}  & \textbf{0.329} &  \textbf{0.45} & \textbf{0.927} & \textbf{0.796} & \textbf{0.742} & \textbf{0.597} \\
            \bottomrule
        \end{tabular}
    }
     \caption{Quantitative results with more advanced base models, i.e., SDXL and kv2.2 \cite{razzhigaev2023kandinsky}.}
    \label{Quantitative_results__2}
\end{table}

\subsection{Main Results}
In this section, We will compare \textit{CustomContrast} with previous SOTAs through qualitative and quantitative analyses.

\textbf{Quantitative results.} 
\textit{CustomContrast} is compared with previous SOTAs on both SD-v1.5 and SDXL. Details of SOTAs are in Appendix C.1. As shown in Tab. \ref{Quantitative_results_1}, our model demonstrates significant improvements in both text controllability and image similarity compared to previous SD-V1.5 methods, with a notable increase in the ImageReward metric from 0.157 to 0.454. The quantitative comparison for SDXL, as shown in Tab. \ref{Quantitative_results__2}, indicates that our method also improves both controllability and similarity, with the CLIP-T  increasing from 0.312 to 0.329.

%

\begin{figure*}[ht]
\begin{center}
\includegraphics[width=\textwidth]{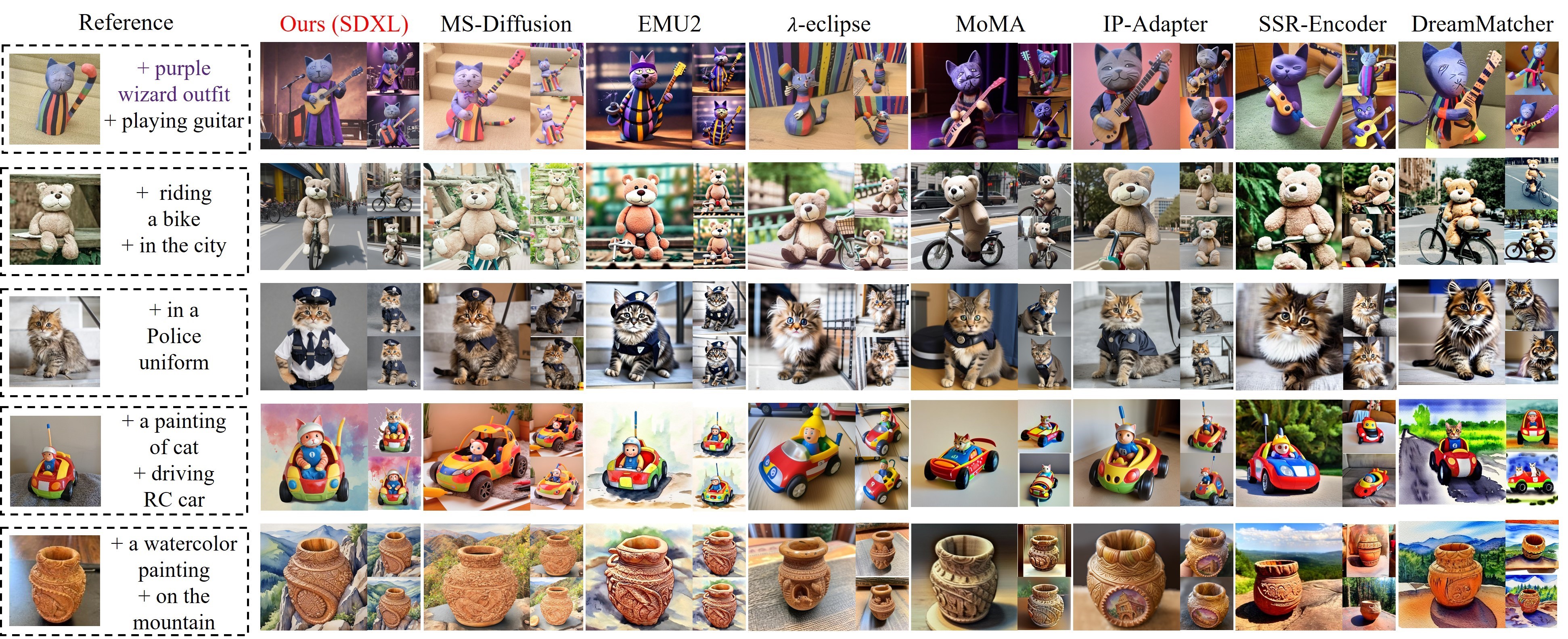}
\end{center}
   \caption{\textbf{Qualitative comparison} with existing methods. \textit{CustomContrast} decouples intrinsic features from redundant features, enabling flexible text control over complex pose (\emph{e.g.}, the cat toy in the first row) and shape (\emph{e.g.}, cat driving car in the fourth row) transformations. In contrast, other methods underperform due to the influence of coupled redundant features.}
\label{fig:example}
\end{figure*}

\textbf{Qualitative results.} Fig. \ref{fig:example}  illustrates the qualitative results compared with existing methods. \textit{CustomContrat} (SDXL) demonstrates high fidelity and controllability, achieving superior zero-shot customization capabilities even for complex accessorization, shape,  and style modification, where other models underperform (\emph{e.g.}, a cat toy playing the guitar in the first row, a driver turning into a cat in the fourth row, and watercolor style in the fifth row).  Qualitative results of \textit{CustomContrast} (SD-v1.5) are in Appendix C.3.


\begin{figure}[!t]
\centering
\includegraphics[width=1\linewidth]{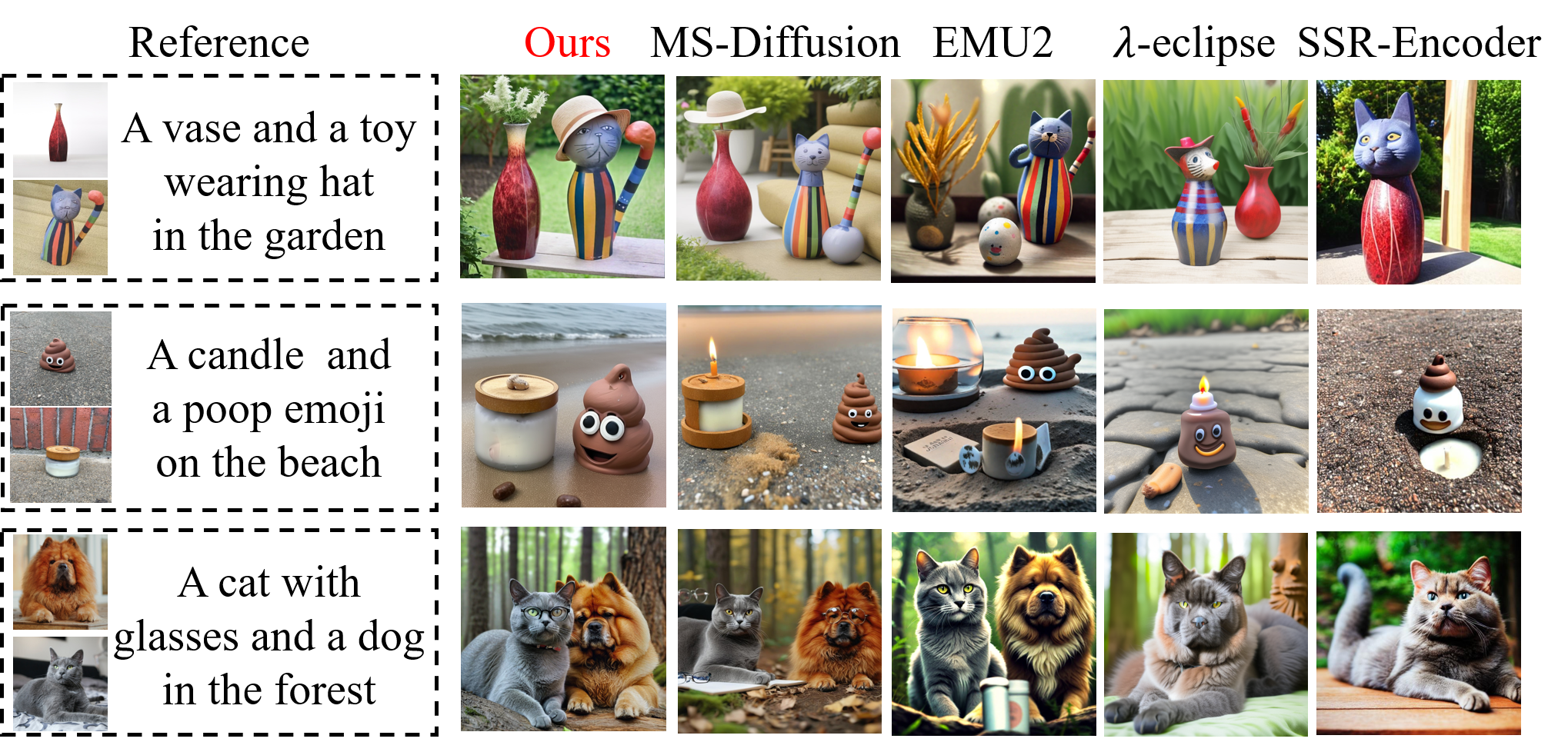}
\caption{\textbf{Qualitative results} of multi-subject generation.}
\label{fig_sdxl_multi_subject}
\end{figure}

To further demonstrate \textit{CustomContrast}'s capabilities in complex editing, we present qualitative results in multi-object and human domain generation. The qualitative results for the multi-subject task are shown in Fig. \ref{fig_sdxl_multi_subject}.  Our model can preserve the multiple subject similarities while achieving superior editing capabilities (\textit{e.g.}, only the cat toy wears a hat in the first row; only the cat wears glasses in the third row). Human domain qualitative results are in Appendix C.5.





\begin{table}[!t]
     \resizebox{8.5cm}{!}
     {
        \begin{tabular}{c c c c c c c}
            \toprule
             ID & Visual & Textual & TV Fusion  &E-CI  $\uparrow$ & E-DI$\uparrow$  & CT$\uparrow$ \\
            \midrule
            0 &  \checkmark & &  &   0.731 &  0.554 &    0.302 \\
            1 &  \checkmark & \checkmark &   & \textbf{0.765} & \textbf{ 0.568}  & 0.282  \\
            2 & \checkmark & \checkmark  &  \checkmark & 0.754& 0.563  & \textbf{0.311}  \\

            \bottomrule
            
        \end{tabular}}
    \caption{Ablation studies on MFI-Encoder. "Visual" and "Textual" refer to Qformer types. Compared with ID-0, the MFI-encoder (ID-2) improves similarity and controllability.}
\label{ablation_0}
\end{table}

\begin{table}[tbp]
     \resizebox{8.5cm}{!}
     {
        \begin{tabular}{c c c c c c c}
            \toprule
             ID & MFI-encoder & CSCL &MACL &E-CI  $\uparrow$ &E-DI$\uparrow$  & CT$\uparrow$ \\
            \midrule
            2 &  \checkmark & &  &  0.754& 0.563  &    0.311  \\
            3 &  \checkmark & \checkmark &   & 0.765 & 0.575  &  0.323 \\
            4 &   \checkmark &  &  \checkmark & 0.780 & 0.584 & 0.318 \\
            5 & \checkmark & \checkmark  &  \checkmark & \textbf{0.788}  & \textbf{0.591}  & \textbf{0.325}  \\

            \bottomrule
            
        \end{tabular}}
    \caption{Ablation studies on the effectiveness of each module based on SD-v1.5.  With both MACL and CSCL losses combined (ID-5),  \textit{CustomContrast} achieves dual optimization in text controllability and image similarity.}
\label{ablation}
\end{table}

\subsection{Ablation Studies}

To demonstrate the effectiveness of essential components of \textit{CustomContrast}, we conduct extensive ablation experiments. All ablation experiments are trained on SD-v1.5. We present the key ablation study in the main text, with the remaining ablation experiments shown in Appendix C.6. 

\textbf{Effectiveness of Each Module for MFI-Encoder.} As indicated in Tab. \ref{ablation_0},  compared to using only Visual-Qformer (ID-0), introducing Textual-Qformer (ID-1) further map the subject features to the $S^+$ space, enhancing subject similarity. However, simply adding Textual-Qformer may reduce text controllability, due to the lack of unified constraints between the visual and textual spaces. To address this issue, we introduce the TV-Fusion module (ID-2), which extracts text-related visual features, simultaneously improving both text controllability and image similarity compared to ID-0.

\textbf{Effectiveness of Contrastive Learning Paradigm.}  As indicated in Tab. \ref{ablation}, the contrastive learning paradigm achieves dual optimization in text controllability and image similarity. 
Compared to the baseline without contrastive learning constraints (ID-2), applying only CSCL (ID-3) significantly improves text controllability. MACL (ID-4) makes the model focus on the subject essential features, thereby achieving better similarity. Meanwhile, due to the decoupling of irrelevant features, the model can better respond to the text. Ultimately, with both losses (ID-5),  \textit{CustomContrast} achieves dual optimization in controllability and similarity, increasing text controllability from 0.311 to 0.325.

\subsection{Visualization}
\begin{figure}[!t]
\centering
\includegraphics[width=1\linewidth]{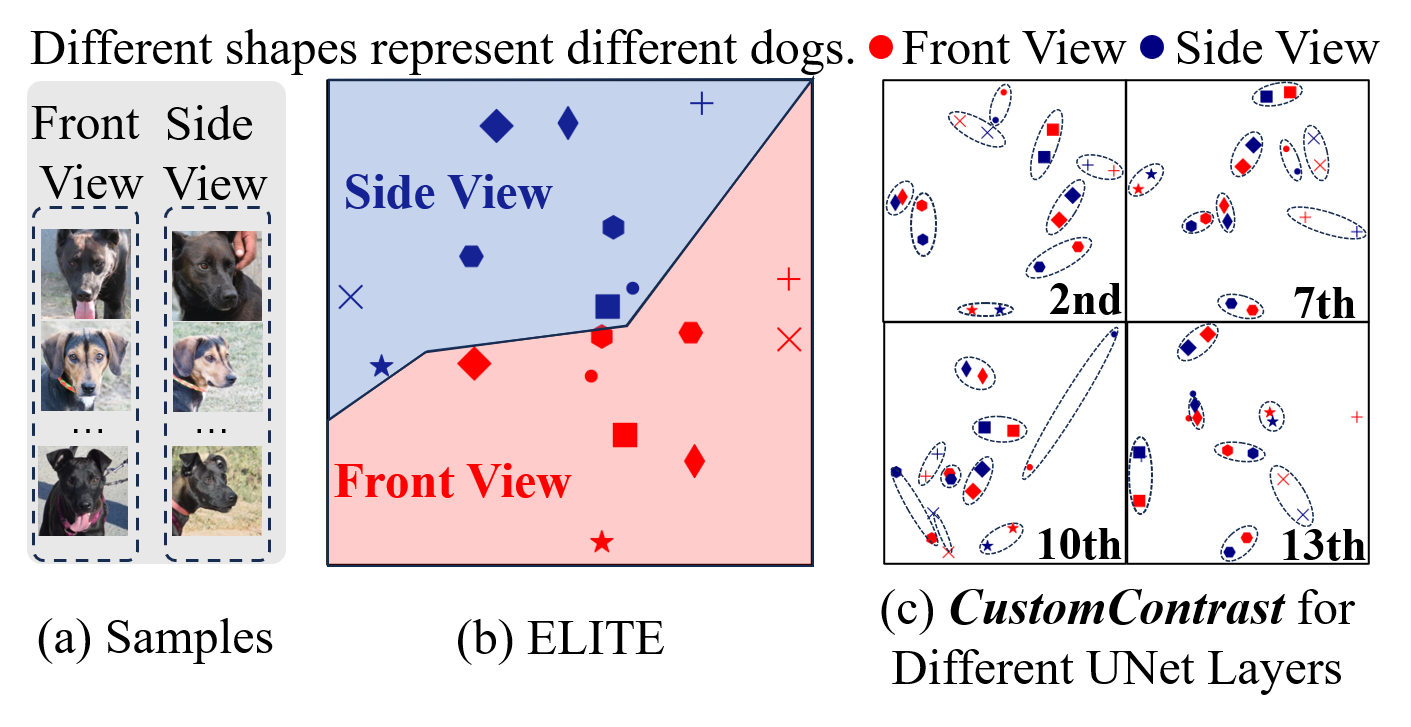}
\caption{(a) Dog's view examples. (b) T-SNE of ELITE representations. (c)  T-SNE of our representations, where 2nd represents the second cross attention, and so on. Ideal representations should achieve  intra-consistency (\emph{i.e.}, points with the same shape should be spatially closer in the figure). ELITE is influenced by different views,  with view features coupled into the representations.  However,  \textit{CustomContrast}  decouples view features from the intrinsic features,  preserving relative distances between samples.}
\label{visualization_1}
\end{figure}



To further illustrate our model's superiority, we use t-SNE to analyze the subject representations of ELITE  \cite{wei2023elite} and our model in Fig. \ref{visualization_1}.  We use dogs with front and side views from DogFaceNet \cite{dogfacenet} as samples. Different shapes represent different dogs, and different colors indicate different views. In Fig. \ref{visualization_1}(b), the ELITE representations are more entangled by view interference. In contrast, $\boldsymbol{s}^+$  learned by \textit{CustomContrast} better preserves relative distances between samples in Fig.  \ref{visualization_1}(c), with different views of the same dog being closer. These analyses demonstrate our model decouples view redundant features from intrinsic subject features.

\section{Conclusion}
In this paper, we propose \textit{CustomContrast}, a novel paradigm for customized T2I from a cross-differential perspective. \textit{CustomContrast} introduces a novel Multilevel Contrastive Learning paradigm that captures intrinsic subject representations from high-level semantics to low-level appearance, supported by the MFI-encoder for consistent cross-modal representations. Extensive experiments demonstrate the superiority of our model, effectively achieving high subject similarity and textual controllability, while easily extending to multi-subject and the human domain generation.
%

\section*{Acknowledgements}
This research is supported by Artificial Intelligence-National Science and Technology Major Project 2023ZD0121200, National Natural Science Foundation of China under Grant 62222212 and National Natural Science Foundation of China under Grant 623B2094.

\bibliography{aaai25}

\end{document}